\documentclass[letterpaper]{article} % DO NOT CHANGE THIS
\usepackage{aaai2026}  % DO NOT CHANGE THIS
\usepackage{times}  % DO NOT CHANGE THIS
\usepackage{helvet}  % DO NOT CHANGE THIS
\usepackage{courier}  % DO NOT CHANGE THIS
\usepackage[hyphens]{url}  % DO NOT CHANGE THIS
\usepackage{graphicx} % DO NOT CHANGE THIS
\usepackage{natbib}  % DO NOT CHANGE THIS AND DO NOT ADD ANY OPTIONS TO IT
\usepackage{caption} % DO NOT CHANGE THIS AND DO NOT ADD ANY OPTIONS TO IT
\usepackage{algorithm}
\usepackage{algorithmic}

\usepackage{newfloat}
\usepackage{listings}
\DeclareCaptionStyle{ruled}{labelfont=normalfont,labelsep=colon,strut=off} % DO NOT CHANGE THIS
\floatstyle{ruled}
\newfloat{listing}{tb}{lst}{}
\floatname{listing}{Listing}
\usepackage{amssymb}
\usepackage{amsmath}
\usepackage{subcaption}

\newtheorem{theorem}{Theorem} % [section] 表示按节编号
\newtheorem{remark}[theorem]{Remark}
\newtheorem{definition}[theorem]{Definition}
\newtheorem{proposition}[theorem]{Proposition}
\newtheorem{assumption}[theorem]{Assumption}
\nocopyright

\title{Controllable Flow Matching for Online Reinforcement Learning}
\author{
    Bin Wang\textsuperscript{\rm 1,\rm 2},
    Boxiang Tao\textsuperscript{\rm 1,\rm 2},
    Haifeng Jing\textsuperscript{\rm 3,\rm 1},
    Hongbo Dou\textsuperscript{\rm 1,\rm 2},
    Zijian Wang\textsuperscript{\rm 1,\rm 2}
}
\affiliations{
    \textsuperscript{\rm 1}College of Computer Science and Technology \& Qingdao Institute of Software, China University of Petroleum (East China), China\\
    \textsuperscript{\rm 2}Shandong Key Laboratory of Intelligent Oil \& Gas Industrial Software, China\\
    \textsuperscript{\rm 3}School of Software \& Microelectronics, Peking University, China\\
}

\begin{document}

\maketitle

\begin{abstract}
Model-based reinforcement learning (MBRL) typically relies on modeling environment dynamics for data efficiency. However, due to the accumulation of model errors over long-horizon rollouts, such methods often face challenges in maintaining modeling stability. To address this, we propose \textbf{CtrlFlow}, a trajectory-level synthetic method using conditional flow matching (CFM), which directly modeling the distribution of trajectories from initial states to high-return terminal states without explicitly modeling the environment transition function. Our method ensures optimal trajectory sampling by minimizing the control energy governed by the non-linear Controllability Gramian Matrix, while the generated diverse trajectory data significantly enhances the robustness and cross-task generalization of policy learning. In online settings, CtrlFlow demonstrates the better performance on common MuJoCo benchmark tasks than dynamics models and achieves superior sample efficiency compared to standard MBRL methods.
\end{abstract}

% Uncomment the following to link to your code, datasets, an extended version or similar.
% You must keep this block between (not within) the abstract and the main body of the paper.
% \begin{links}
%     \link{Code}{https://aaai.org/example/code}
%     \link{Datasets}{https://aaai.org/example/datasets}
%     \link{Extended version}{https://aaai.org/example/extended-version}
% \end{links}

\section{Introduction}
In recent years, deep reinforcement learning (DRL) has shown remarkable decision-making capabilities in complex tasks \cite{walk:25,DRL1,DRL2,DRL3,DRL4} through a \textit{trial-and-error} process. However, particularly in online RL settings, data efficiency has emerged as a critical bottleneck hindering practical applications. A widely adopted approach is MBRL \cite{MBPO,InformedPOMDP,InfoProp,MBCSND,LEQ-25}, which learns environment dynamics models (including the state transition and reward function) to generate simulated interactions and enable policy training with significantly reduced real-world data sampling.

Nevertheless, such methods are highly dependent on the accuracy of the learned environment dynamics modeling. A well-known limitation of learned models is their tendency to accumulate errors during long-horizon rollouts, resulting in synthetic trajectories that diverge from the real state-action distribution, ultimately degrading policy optimality. Consequently, enhancing the quality of generated data while keeping sample efficiency rises as the key to policy optimization.

For tackling the notorious cumulative error problem in MBRL, existing research focuses on improving model accuracy, quantifying uncertainty, and planning robustness. One key approach involves using ensemble dynamics models \cite{MBPO,ME-TRPO,STEVE,PETS,InfoProp} to average predictions and estimate epistemic uncertainty, while probabilistic neural networks can capture aleatoric uncertainty for better error bounds. In addition, some methods \cite{ccem,MPPVE,ADMPO} use multi-step planning strategies to reduce prediction frequency, thereby reducing error accumulation. However, these methods only set upper bounds for errors or mitigate deviations in dynamic simulations by reducing bootstrapping in timesteps, and do not truly solve the problem.

\begin{figure}[t]
    \centering
    \includegraphics[width=1.0\linewidth]{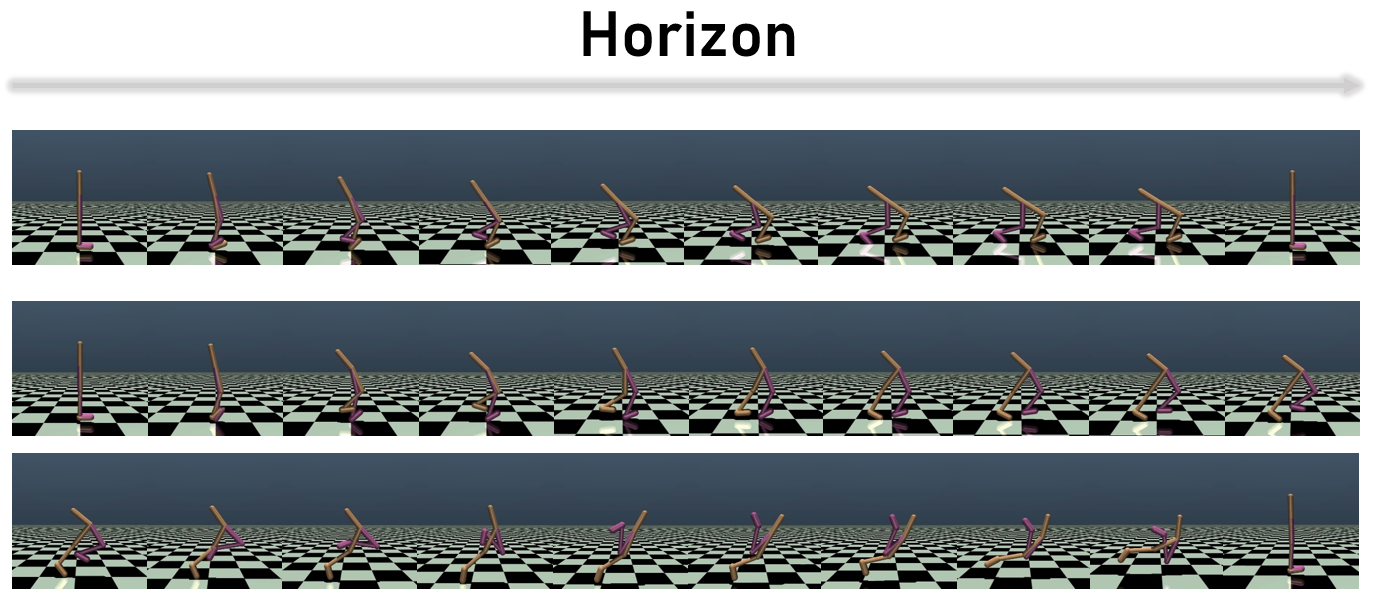}
    \caption{A validation episode in the Walker2d benchmark task of MuJoCo. The Soft Actor-Critic (SAC) algorithm is employed to optimize the policy in the same seed. In the first row (\textbf{Failed} \includegraphics[width=0.3cm]{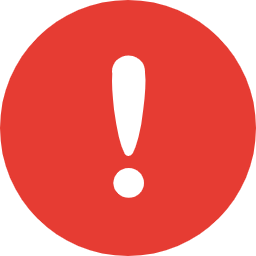} \textbf{Reward: 23} \includegraphics[width=0.3cm]{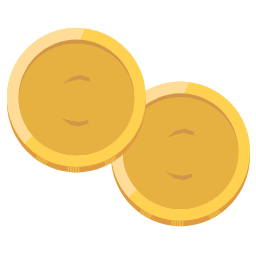}), the trajectory data directly generated by the conditional flow matching model is used for policy training. In the second and third rows (\textbf{Success} \includegraphics[width=0.3cm]{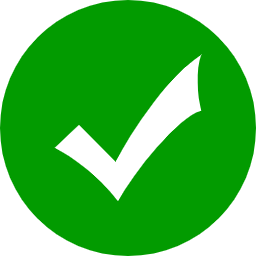} \textbf{Reward: 392} \includegraphics[width=0.3cm]{images/reward.png}), on the basis of the flow matching model, data with control correction and value guidance is utilized.}
    \label{fig:walker2d_episode}
\end{figure}

To fundamentally address this limitation, we note that generative models have demonstrated remarkable success across diverse domains, while such models have also emerged as a powerful tool in reinforcement learning (RL) \cite{RL}, enhancing trajectory generation, planning, and policy learning \cite{Dacer}. These methods leverage diffusion's \cite{Sohl-Dickstein_Weiss_Maheswaranathan_Ganguli_2015,Ho_Jain_Abbeel_2020} iterative denoising process to generate high-quality \cite{Diffuser}, diverse behaviors while addressing key challenges in RL, such as long-horizon reasoning and non-Gaussian policy distribution. 
Some works \cite{HDiffuser} incorporate hierarchical planning to improve scalability, allowing the model to efficiently reason over extended time horizons. Others \cite{AdaptiveDiffuser} enhance generalization by combining diffusion with evolutionary optimization, adapting to unseen tasks through iterative refinement. Additionally, diffusion has been reinterpreted as a reverse RL process \cite{Merlin}, where policy learning aligns with the denoising trajectory. However, these methods have been predominantly applied to offline RL, while the potential of generative models in online settings remains underexplored.

Thus, from the motivation of avoiding the curse of this kind of bias from step-by-step prediction, we propose a novel approach \textbf{C}on\textbf{tr}o\textbf{l}lable \textbf{Flow} matching (\textbf{CtrlFlow}) for high-quality trajectory data synthesis in online RL. Our method transforms Conditional Vector Fields (CVFs) \cite{lipman2022flow} into value guidance energy-based \cite{mei,egds} vector fields \cite{feng2025guidance,zhang2025energy} through shrinkage in return scaling, enabling trajectory sampling can be aligned towards high-return regions. While, we introduce non-linear Controllability Gramian Matrix to augment the model's capacity for dynamic trajectory distribution modeling. This design preserves global controllability from arbitrary initial distributions to target trajectories during sampling. It should be emphasized that our approach is fundamentally not MBRL in nature, as its operational mechanism does not involve explicit simulation of environmental dynamics. Nevertheless, the proposed method can functionally substitute the dynamics model in MBRL frameworks, achieving effective data augmentation through its unique generative paradigm.

Figure \ref{fig:walker2d_episode} reveals that CtrlFlow exhibits remarkable performance in synthesizing trajectory data, significantly enhancing the policy learning process through its generation mechanism. The main contributions are three-fold. Firstly, to the best of our knowledge, this is the first work adapts flow matching method for online RL that works on the trajectory-level data generation, effectively solve the inherent error accumulation problem in MBRL, rather than merely mitigating it. Secondly, our approach significantly enhances the model's adaptability to Markovian trajectory distributions while ensuring efficient and stable generation through the novel introduction of a Controllability Gramian Matrix to reduce control energy. Lastly, we design an energy-based trajectory optimization method in online RL that achieves aligned generation of high-return trajectories through guidance vector fields. We evaluate CtrlFlow on MuJoCo benchmark tasks and all experiments followed the Dyna-style paradigm. The results demonstrate that our method exhibits superior performance and better convergence than compared MBRL approaches.

\section{Related Work}
\subsection{Normalizing Flows and Flow Matching Models}
Normalizing flows are a class of generative models that construct complex probability distributions through a sequence of invertible and differentiable transformations applied to simple base distribution, typically a multivariate Gaussian. To scale up the training of normalizing flows in continuous setting, recent works \cite{albergo2022building,lipman2022flow,liu2022flow} propose an effective simulation-free approach by parameterizing the vector field that flows noise samples to data samples. \cite{lipman2022flow} propose Conditional Flow Matching (CFM), an extension of Flow Matching (FM), to train continuous normalizing flows by mixing simpler conditional probability paths. \cite{yang2024consistency} further propose Consistency Flow Matching to enable FM method more stable and self-consistency in the velocity field. FM models offer a more efficient and stable alternative to diffusion models for RL trajectory generation by eliminating iterative denoising steps and enabling direct probability path optimization with deterministic dynamics.

\subsection{Flow Matching in Decision Making and RL}
In decision making and RL, the FM model enables efficient learning of policies and value functions by interpolating between initial and target state distributions. \cite{zheng2023guided} apply the integration of guidance and flow-based models for return-conditioned plan generation in offline RL enables this planning to generalize to unseen conditional signals during evaluation. QIPO \cite{zhang2025energy} propose a energy-guide velocity field based on the conditional velocity field to generate the energy-guided distribution without any auxiliary model. FQL \cite{park2025flow} further extend the CFM model to Q-learning, it involves highly multimodal action distributions on complex tasks and more suitable for policy learning in offline RL methods, without requiring iterative flow steps at test time. FlowQ \cite{alles2025flowq} scales the training time via the number of flow generation steps without backpropagate gradients through actions sampled from the flow policy. However, there is no flow matching variant has yet been successfully deployed in online RL to handle the compounding error problem in dynamics model learning.

\section{Preliminaries}
In this section, we introduce the fundamental concepts of Reinforcement Learning, Conditional Flow Matching models, Controllability Gramian Matrix in particular.

\subsection{Problem Setup}
Reinforcement Learning (RL) is always formalized as a Markov Decision Process (MDP), defined by a tuple $(\mathcal{S},\mathcal{A},P,r,\gamma)$, where $\mathcal{S}\in\mathbb{R}^{d_s}$ is the state space, $\mathcal{A}\in\mathbb{R}^{d_a}$ is the action space, $P(s_{h+1}\mid s_h,a_h)\in\mathcal{S}\times\mathcal{A}\rightarrow\mathcal{S}$ specifies the state transition probability which is conditioned by state $s_h\in\mathcal{S}$ and action $a_h\in\mathcal{A}$ together at time $h$. $r=\mathcal{R}(s,a)\in\mathbb{R}$ is the reward function and $r_h$ is the immediate reward at time $h$, $\gamma\in [0,1)$ is the discounted factor. A policy $\pi$ is often randomly initialized from an initial state $\rho_0$, and then, the agent learns $\pi(a_h\mid s_h)$ through interactions with the environment, aiming to maximize the expected long-term cumulative return: $\mathbb{E}_\pi\left[\sum^\infty_{h=0}\gamma^h\mathcal{R}(s_h,a_h)\right]$.

\subsection{Conditional Flow Matching Models}
Let $\mathbb{R}^d$ denote the data space with data points $x^t\in\mathbb{R}^d$\footnote{In this paper, the superscript time denotes the model's internal timestep, while the subscript time represents timestep in RL horizon.}. Here the subscript $t\in[0,1]$ represents inference time in generation models. At time $t=0$, the data $x^0$ is sampled from a simple distribution $p(x^0)=\mathcal{N}(x^0\mid 0,I)$. And at time $t=1$, the data $\tau^1$ is generated from a complex distribution $p(x^1)$. Continuous Normalizing Flows has two important objects in it: the \textit{probability density path} $p^t:[0,1]\times\mathbb{R}^d\rightarrow\mathbb{R}_{>0}$ which follows $\int p^t(x)dx=1$, and a \textit{time-dependent vector field}, $v^t:[0,1]\times\mathbb{R}^d\rightarrow\mathbb{R}^d$. The \textit{flow} $\Phi:[0,1]\times\mathbb{R}^d\rightarrow\mathbb{R}^d$ is constructed by the vector field $v$ via the ordinary differential equation (ODE): $\frac{d}{dt}\Phi^t(x)=v^t(\Phi^t(x))$, where $\Phi^0(x)=x$.

Given a target probability density path $p^t(x)$ which is generated by its corresponding vector field $\widehat{v}^t(x)$, the Flow Matching (FM) objective as: $\mathcal{L}_\text{FM}(\theta)=\mathbb{E}_{t,p^t(x)}\|v^t(x)-\widehat{v}^t(x)\|^2$, where $v_\theta^t$ is a neural network with learnable parameter $\theta$. However, $p^t$ and $\widehat{v}^t$ are always intractable to use in practice. To address this issue, \textit{conditional probability path} $p^t(x^t\mid x^1)$ has been suggested to be generated via \textit{conditional vector field} $\widehat{v}^t(x^t\mid x^1)$, it provides a simple way to model the marginal probability path $p^t(x)$. In practice, the construction of $p^t(x^t\mid x^1)$ and $\widehat{v}^t(x^t\mid x^1)$ is always chosen Gaussian distribution, i.e., $p^t(x^t\mid x^1)=\mathcal{N}(x^t\mid \mu^t(x^1),\sigma^t(x^1)^2I)$, where $\mu: [0,1]\times\mathbb{R}^d\rightarrow\mathbb{R}^d$ is the time-dependent mean of the Gaussian distribution, while $\sigma: [0,1]\times\mathbb{R}\rightarrow\mathbb{R}_{>0}$ is standard deviation.

\subsection{Controllability Gramian Matrix}
The Controllability Gramian Matrix is a fundamental concept in control theory that quantifies the controllability of a dynamical system. For a linear time-invariant (LTI) system described by: $\dot{x}^t=Ax^t+Bu^t$, the Controllability Gramian Matrix $W_c$ over a time horizon $t\in[0,T]$ is defined as:
\begin{equation}\label{eq1}
    W_c:=\int^T_0 e^{At}BB^\top e^{A^\top t}dt,
\end{equation}
where $A(t,x)\in\mathbb{R}^n\times\mathbb{R}^n$ is a matrix which governs the intrinsic dynamics of the system state $x^t$ in the absence of control inputs: $\dot{x}^t=Ax^t$ (\textit{unforced system}). And matrix $B(t,x)\in\mathbb{R}^n\times\mathbb{R}^m$ maps control actions $u^t$ to state derivatives: $\dot{x}^t=Bu^t$ (\textit{forced system}), where $m$ is control input dimension. The Gramian characterizes the minimum control energy required to drive the system from the origin to any target state $x^T$:
\begin{equation}\label{eq2}
    \min_u\int^T_0\|u^t\|^2dt={x^0}^\top W_c^{-1}x^0.
\end{equation}

In this work, we consider that lower control energy requirements of a system directly correlate with reduced susceptibility to noise perturbations during the sampling process.
Unless otherwise specified, we are interested in non-linear control-affine system:
\begin{equation}\label{eq3}
    \dot{x}^t=v(t,x^t)+B(t,x^t)u^t,\quad x^{t_0}=x^0,\quad t\in[0,1],
\end{equation}
where $A(t,x)=\text{Id}$ and $v^t$ is a nonautonomous, non-linear vector field which can be constructed by a neural network.

\noindent \textbf{Notation}\ \ \ Let $\widehat{\tau}$ represents generated trajectories by the model and $\tau$ denotes trajectories sampled from replay buffer.

\section{Method}
In this section, we propose a CFM model as a replacement for environment dynamics models in MBRL. The Figure \ref{fig:overall} shows the overall of our approach and see detail algorithms in Appendix B.

\subsection{Trajectory-Level Flow Matching Model}\label{sub31}
To avoid the accumulation of errors over time caused by step-by-step prediction using environment dynamics models, we suggest generating entire trajectories directly instead of bootstrapping each state transition sequentially. This approach helps preserve the distributional consistency of the trajectories, enabling the generated data that more closely resembles that from the real environment, thereby improving the stability and efficiency of policy optimization. To this end, we introduce a CFM model into the data augmentation in online settings and define a trajectory-level generative model, as shown in Definition \ref{define1}:
\begin{definition}[Trajectory-Level Flow Matching Model]\label{define1}
    Assume that the trajectory generated by policy $\pi_\xi$ in the RL environment is denoted as $\tau=\{(s_1,a_1,r_1),(s_2,a_2,r_2),\cdots,(s_H,a_H,r_H)\}$, where $s_h\in\mathcal{S}$, $a_h\in\mathcal{A}$, $r_h\in\mathbb{R}$, and $H$ is the maximum trajectory length. The model $\mathcal{M}$ aims to learn a parameterized continuous vector field $v_\theta$ to model and sample from the trajectory distribution $p_\pi(\tau)$, conditioned on the current policy $\pi_\xi$.
\end{definition}

In order to ensure that the trajectories sampled from the model can preserve the environmental dynamics $P=\{s_{h + 1}, r_{h + 1}\mid s_h, a_h\}$ ($1\leq h\leq H$), the model employs Transformer \cite{attention} to model the trajectories and incorporates a temporal encoding mechanism to capture the relative relationships among different time steps in the trajectories. This design strengthens the information representation of the temporal structure, which is particularly crucial for improving the prediction accuracy of trajectory endpoints.

The model $\mathcal{M}$ can generate trajectories of arbitrary lengths ($1\leq h\leq H$), thereby enhancing the ability to align distribution and the efficiency of modeling data diversity. Although $\mathcal{M}$ can avoid distribution shifts caused by cumulative errors in the direction of trajectory time steps $h$, errors may still occur between iterative time steps $t$ during the generation process. To address this issue, we will introduce the corresponding solutions in the next subsection.

\begin{algorithm}[tb]
\caption{Training CFM model $\mathcal{M}_\theta$}
\label{alg:1}
\textbf{Input}: $H$, $\mathcal{M}_\theta$, $\mathcal{B}_\text{env}$ after warming up, variance $\sigma$, policy $\pi_\xi$\\
% \textbf{Parameter}: Optional list of parameters\\
\textbf{Output}: CFM model $\mathcal{M}_\theta$
\begin{algorithmic}[1] %[1] enables line numbers
\FOR{e epochs}
\STATE Randomly sample an integer $h$ from $[1, H]$ uniformly
\STATE Sample $\tau^1=\{(s_1,a_1,r_1),(s_2,a_2,r_2),(s_3,a_3,r_3)\dots$ $,(s_h,a_h,r_h)\}$ from $\mathcal{B}_\text{env}$
\STATE Randomly sample $\widehat{\tau}^0\sim\mathcal{N}(\widehat{\tau}^0\mid 0,I)$
\STATE Randomly sample time $t\in[0,1]$ uniformly, and noise $\epsilon^t$
\STATE Mean of distribution $\mu^t=t\cdot\tau^1+(1-t)*\widehat{\tau}^0$ , construct conditional probability density path $\widehat{\tau}^t=\mu^t+\sigma^t\cdot\epsilon^t$
\STATE Target Conditional Vector Field $\widehat{v}=\tau^1-\widehat{\tau}^0$
\STATE Minimize $\mathcal{L}_\text{CFM}$ defined in Equation \ref{eq5}
\ENDFOR
\end{algorithmic}
\end{algorithm}

\begin{figure*}[t]
\centering
\includegraphics[width=1.0\textwidth]{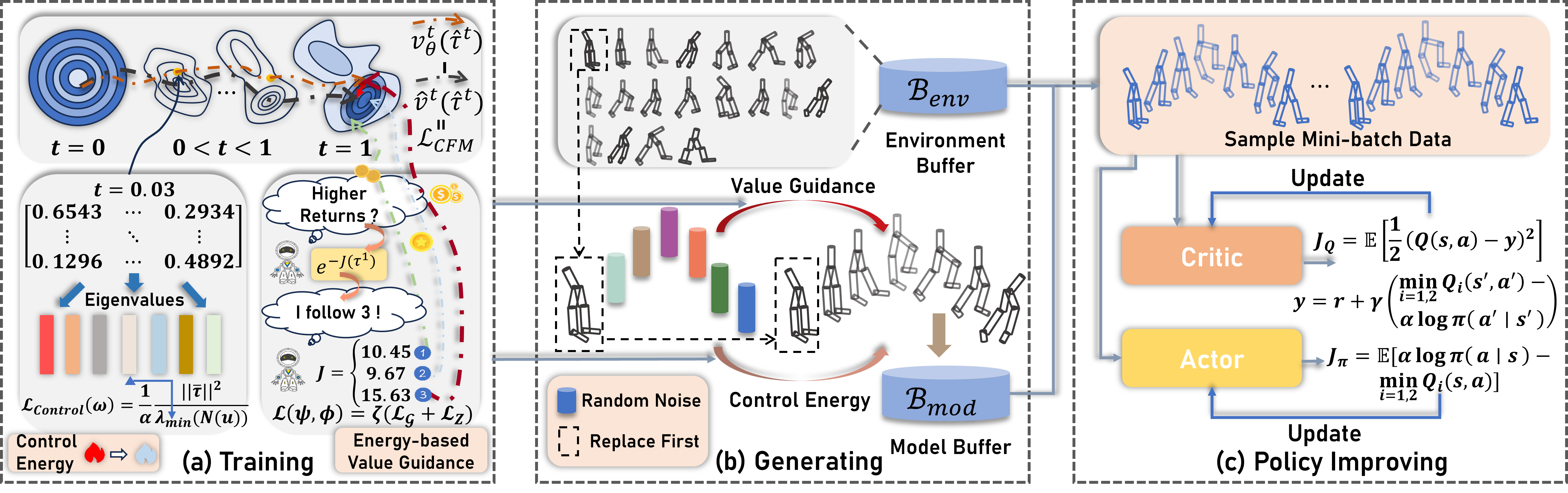} % Reduce the figure size so that it is slightly narrower than the column.
\caption{The overall of CtrlFlow.}
\label{fig:overall}
\end{figure*}

\begin{definition}[The Input of Model $\mathcal{M}$]\label{define33}
    At iterative time step $t=0$, the generation length of the trajectory $\widehat{\tau}$ is chosen as $h\in[1,H]$, and the trajectory data $\widehat{\tau}^0$ is randomly initialized as the input to the model:
    \begin{equation}\label{eq4}
        \widehat{\tau}^0:=\begin{pmatrix}
                            s_1,s_2,s_3,\cdots,s_h\\
                            a_1,a_2,a_3,\cdots,a_h\\
                            r_1,r_2,r_3,\cdots,r_h\\
                            \end{pmatrix}
    \end{equation}
\end{definition}

Given a joint transition distribution of states, actions, and rewards as the input of $\mathcal{M}$ in Definition \ref{define33}. In practice, the initial trajectory distribution follows a standard Gaussian distribution, i.e., $p(\widehat{\tau}^0)=\mathcal{N}(\widehat{\tau}^0\mid 0,I)$. By sampling real trajectory data $\tau^1$ of the same length $h$ from the environment replay buffer $\mathcal{B}_\text{env}$, a conditional probability path distribution $\widehat{\tau}^t$ between $\widehat{\tau}^0$  and $\tau^1$ is constructed using linear interpolation to characterize the iterative process from $t=0$ to $t=1$. The goal of the model is to learn a parameterized vector field $v_\theta$ that can describe the velocity of the particle at position $\widehat{\tau}^t$. Therefore, the training objective is minimizing the following conditional flow matching loss function:
\begin{equation}\label{eq5}
\begin{aligned}
    \mathcal{L}_\text{CFM}(\theta)=
    \mathbb{E}_{\tau^1\sim \mathcal{B}_{\text{env}},p(\widehat{\tau}^t\mid\tau^1),t\in[0,1]}\\
    [\|v_\theta(\widehat{\tau}^t,t)-\widehat{v}(\widehat{\tau}^t,t\mid \tau^1)\|^2],
\end{aligned}
\end{equation}
where $\widehat{v}(\widehat{\tau}^t,t\mid \tau^1)$ represents the analytical expression of the matching direction for the trajectory sample from $\widehat{\tau}^0$ to the target trajectory data $\tau^1$ at time $t\in[0,1]$. The complete training process of the CFM model $\mathcal{M}$ is described in Algorithm \ref{alg:1}. 

To stabilize the training process, we introduce an additional KL divergence term following $\mathcal{L}_\text{CFM}$ to further regularize the probability distribution between the learned vector field $v_\theta$ and its target $\widehat{v}$ at time $t=1$. Experimental results demonstrate that this approach improves distribution alignment, thereby enhancing generation quality and accelerating training convergence.

\subsection{Controllable Flow for Distribution Adaptation}\label{sub32}
In the before subsection, the CFM model was proposed to solve the cumulative error problem in trajectory generation. However, learning an adaptive vector field remains challenging due to the dynamically shifting data distribution in online settings. Consequently, we reformulate CNFs as a series of affine transformations and introduce control theory \cite{tamekue2025control,mei2024flow}, where the designed control data generation enables optimal control energy from initial states to target states.

Recall the non-linear control system in equation \ref{eq3}, within this context, the vector field $v$ characterizes the intrinsic non-linear dynamics of the system, i.e., sampling process, serving as the core learning objective in the CFM model $\mathcal{M}$. Moreover, we will discuss that the design of control input $u$ and it's regulatory on the vector field $v$. The objective of $u$ is to enhance the generalization capability of $\mathcal{M}$ across dynamic distribution and improve the robustness of generated data, achieved through minimizing control energy constraints.

\begin{assumption}
    The vector field $v^t: [0,1]\times\mathbb{R}^d\rightarrow\mathbb{R}^d$ satisfies the following assumptions
    \begin{enumerate}
        \item The map $t\mapsto v^t(\widehat{\tau})$ belongs to $L^\infty(0,1)$ , for every fixed $\widehat{\tau}\in\mathbb{R}^d$.
        \item The map $\widehat{\tau}\mapsto v(\widehat{\tau}^t)$ is $C^2$, for every fixed $t\in[0,1]$.
    \end{enumerate}
    Furthermore, at every time $t\in[0,1]$, the vector field $v^t$ is globally $\Lambda$-Lipschitz continuous with $\Lambda>0$.
\end{assumption}

\begin{definition}
    The intrinsic non-linear dynamics are described by the vector field $v$ as a function of time:
    \begin{equation}
        \frac{\partial}{\partial t}\Phi^{t_0,t}(\widehat{\tau}^0)=v(\Phi^{t_0,t}(\widehat{\tau}^0)),\quad \Phi^{0,0}(\widehat{\tau}^0)=\widehat{\tau}^0\in\mathbb{R}^d,
    \end{equation}
    where $\{\Phi^{\mathbf{s},t}\mid (\mathbf{s},t)\in[0,1]^2\}$ is a flow represented as a two-parameter family of diffeomorphisms, satisfying the following algebraic relations:
    \begin{equation}
    \begin{aligned}
        \Phi^{t,t}&=\text{Id},\quad \forall\ t\in[0,1],\\
        \Phi^{t_2,t_3}\circ\Phi^{t_1,t_2}&=\Phi^{t_1,t_3},\quad \forall\ (t_1,t_2,t_3)\in[0,1]^3,\\
        (\Phi^{t_1,t_2})^{-1}&=\Phi^{t_2,t_1},\quad \forall\ (t_1,t_2)\in[0,1]^2.
    \end{aligned}
    \end{equation}
\end{definition}

\begin{theorem}
    For any $\widehat{\tau}^0\in\mathbb{R}^d$ and every $b\in B\in L^\infty$, the solution $\widehat{\tau}$ of $\dot{\widehat{\tau}}^t=v^t(\widehat{\tau}^t)+b^t$, $\widehat{\tau}^{t_0}=\widehat{\tau}^0$ can be expressed in following form:
    \begin{equation}
    \begin{aligned}
        \widehat{\tau}^t=\Phi^{T,t}\left(\Phi^{t_0,T}(\widehat{\tau}^0)+\int^t_{t_0}\mathbf{D}\Phi^{\mathbf{s},T}(\widehat{\tau}^\mathbf{s})b^\mathbf{s}d\mathbf{s}\right),\\
        \quad \forall\ t\in[t_0,T],
    \end{aligned}
    \end{equation}
    where $\mathbf{D}$ denotes the Jacobian matrix of the flow $\Phi$ with respect to time at the initial state $\widehat{\tau}^0$, and is defined for every fixed $\widehat{\tau}^0\in\mathbb{R}^d$. Proof in Appendix A.1.
\end{theorem}

\begin{definition}[Non-linear Controllability Gramian]
    For a fixed initial state $\widehat{\tau}^0\in\mathbb{R}^d$, control input $u$, and time $t\in[t_0,T]$, the Controllability Gramian Matrix of the non-linear system is defined as:
    \begin{equation}\label{eq10}
    \begin{aligned}
        N(u)&:=N(T,t_0,\widehat{\tau}^0,u)\\
        &=\int^T_{t_0} \mathbf{D}\Phi^{t,T}(\widehat{\tau}^t)B^t{B^t}^\top \mathbf{D}\Phi^{t,T}(\widehat{\tau}^t)^\top dt,
    \end{aligned}
    \end{equation}
    where $\widehat{\tau}^t$ denotes the controlled system state under influence of $u$, $B^t$ is the time-varying control matrix, $\mathbf{D}\Phi^{t,T}(\widehat{\tau}^t)$ represents the Jacobian of the flow map from time $t$ to $T$ evaluated at $\widehat{\tau}^t$.
\end{definition}

\begin{remark}
    The generalized Controllability Gramian Matrix $N(u)$ characterizes state-transition properties under control input $u$ through its integral formulation. Crucially, it transforms the local controllability analysis of a non-linear system into time-varying linear system problem via the Jacobian $\mathbf{D}\Phi^{t,T}$ of the flow map.
\end{remark}
The positive definiteness of $N(u)$ guarantees that the system can reach any state from $\widehat{\tau}^0$ under admissible controls, while its eigenvalue spectrum quantifies directional control effort:
\begin{equation}
\begin{aligned}
    \lambda_{\min}(N(u))\parallel\delta \widehat{\tau}\parallel^2&\leq\delta\widehat{\tau}^\top N(u)\delta \widehat{\tau}\\&\leq\lambda_{\max}(N(u))\parallel\delta \widehat{\tau}\parallel^2,
\end{aligned}
\end{equation}
where $\lambda_\text{min}$, $\lambda_\text{max}$ denotes the smallest and the largest eigenvalue of the matrix $N(u)$, respectively. And larger eigenvalues correspond to more controllable directions. In practice, we define $t_0=0$ and $T=1$. Thus, $N(u)$ can be interpreted as representing the controllability of generated data in the sampling process by control inputs over the time interval $t\in[0,1]$. Additionally, the control input matrix $B^t$ is set to modulate the control intensity at different time $t$, which we refer to as the \textit{\textbf{time-varying weight matrix}}.

\begin{algorithm}[tb]
\caption{Training Controllable Guidance Model $v^\omega_\mathcal{C}$}
\label{alg:2}
\textbf{Input}: $v_\mathcal{C}^\omega$, $\mathcal{M}_\theta$ with fixed parameter $\theta$, horizon $h\in[1,H]$\\
% \textbf{Parameter}: Optional list of parameters\\
\textbf{Output}: Model $v_\mathcal{C}^\omega$
\begin{algorithmic}[1] %[1] enables line numbers
\FOR{$t=0$ \TO $t=1$}
\STATE Generated trajectory data effected by $u$ $\widehat{\tau}^t=v_\mathcal{C}^\omega(\tau^t)$
\STATE The Flow $\Phi^t=\mathcal{M}_\theta(\widehat{\tau}^t)$
\STATE Matrix $B^t=\sum^h_{i=1}\gamma_{i}\mathcal{R}(s_i,a_i)\cdot(1-\mathbb{I}(s_i,a_i))$
\STATE Non-linear Controllability Gramian Matrix $N(u)=\Phi^t(\widehat{\tau}^t)B^t{B^t}^\top \Phi^t(\widehat{\tau}^t)^\top$
\ENDFOR
\STATE $\bar{\tau}=\tau^1-\mathcal{M}_\theta(\widehat{\tau}^0)$
\STATE $\mathcal{L}_\text{Control}(\omega)=\frac{1}{\alpha}\frac{\text{pow}(\bar{\tau},2)}{\lambda_\text{min}(N(u))}$
\end{algorithmic}
\end{algorithm}

\begin{theorem}[Control Input $u$]
Given an initial state $\widehat{\tau}^0\in\mathbb{R}^d$ and a terminal state $\tau^1\in\mathbb{R}^d$, the control input $u$ drives the state transition of the system from $t_0$ to $T$. For any control time $t\in[t_0,T]$, the control input can be defined as:
\begin{equation}
    u^t={B^t}^\top \mathbf{D}\Phi^{t,T}(\widehat{\tau}^t)^\top N^{-1}(\tau^1-\Phi^{t_0,T}(\widehat{\tau}^0)),
\end{equation}
where the matrix $N$ is invertible. Proof in Appendix A.2.
\end{theorem}

This theorem provides an analytical expression for the non-linear control input. The core idea is to construct the control law through the inverse of the Controllability Gramian Matrix $N$, enabling precise steering of the system from the initial state $\widehat{\tau}^0$ to the target state $\tau^1$. Specifically, the control coupling term ${B^t}^\top$ captures the time-varying influence of actuators on the system state, while the linearized state transition $\mathbf{D}\Phi^{t,T}$ locally approximates the non-linear dynamics via the Jacobian matrix, ensuring that the control aligns with the system's evolution direction. Finally, the Gramian-based correction term $N^{-1}(\tau^1-\Phi^{t_0,T}(\widehat{\tau}^0))$ maps the state error to an optimal control input.

\begin{proposition}
    Under the assumption that the Controllability Gramian Matrix is invertible, its inverse reflects the magnitude of energy required for control. To ensure global controllability of the system, we constrain the control energy using the eigenvalues of the Gramian matrix:
    \begin{equation}
        \int^T_{t_0}|u^t|^2 dt=|\tau^1-\Phi^{t_0,T}(\widehat{\tau}^0)|^2_{N^{-1}}\leq \frac{|\tau^1-\Phi^{t_0,T}(\widehat{\tau}^0)|^2}{\lambda_\text{min}(N(u))}.
    \end{equation}
\end{proposition}

\begin{definition}[Controllability Vector Field]
    At any time $t\in[0,1]$, given a conditional vector field $v^t$ learned by a CFM model and a control input $u^t$ with $B^t$, the controllability vector field can be defined as:
    \begin{equation}
        v_\mathcal{C}(\widehat{\tau}^t)=v(\widehat{\tau}^t)+B(\widehat{\tau}^t)u(\widehat{\tau}^t),\quad t\in[0,1]
    \end{equation}
    with
    \begin{equation}
        B(\widehat{\tau}^t)=\sum^h_{i=1}\gamma_{i}\mathcal{R}(s_i,a_i)\cdot(1-\mathbb{I}(s_i,a_i)),
    \end{equation}
    where $\mathbb{I}(s_i,a_i)$ denotes the indicator  function with respect to the state and action at timestep $i\in[1,h]$.
\end{definition}

The complete training of controllable guidance model is shown in Algorithm \ref{alg:2}.

\subsection{Value Guidance with Energy Vector Field}
In previous subsection, we propose a trajectory-level data generation method using conditional flow matching model and it ensures adaptation to the data distribution through the Controllability Gramian Matrix. However, since the model only achieve basic alignment with the target data distribution, the generated trajectory data fails to provide effective policy optimization guidance signals. To address this limitation, we need to directionally modify the original vector field to enable generation of higher-quality trajectories with greater cumulative discounted returns. Building on this insight, we introduce a value guidance vector field that transforms the original vector field into an energy-based vector field through an energy function.

\begin{theorem}[Theorem 3.1. in \cite{feng2025guidance}]
    Adding the guidance vector field $\mathcal{G}(\widehat{\tau}^t)$ to the original vector field $v(\widehat{\tau}^t)$ will form vector field $v^\prime(\widehat{\tau}^t)$ that generates $p^\prime(\widehat{\tau}^t)=\int p^t(\widehat{\tau}^t\mid z)dz$, where $z=(\widehat{\tau}^0,\tau^1)$, as long as $\mathcal{G}(\widehat{\tau}^t)$ follows:
    \begin{equation}
        \mathcal{G}(\widehat{\tau}^t)=\int\left(\mathcal{P}\frac{e^{-J(\tau_1)}}{Z_t(\widehat{\tau}^t)}-1\right)v^t_z(\widehat{\tau}^t\mid z)p(z\mid\widehat{\tau}^t)dz
    \end{equation}
    with,
    \begin{equation}
        Z^t(\widehat{\tau}^t)=\int e^{-J(\tau_1)}p(z\mid\widehat{\tau}^t)dz,
    \end{equation}
    where $\mathcal{P}$ is the reverse coupling ratio.
\end{theorem}

\begin{definition}[Energy Vector Field]
    Given an energy function $J:\mathbb{R}^d\rightarrow\mathbb{R}$ and a CFM model $\mathcal{M}_\theta$ for sampling from $p(\tau)$, the energy-based samples follow the distribution: $\widehat{\tau}\sim p^\prime(\tau)=\frac{1}{Z}p(\tau)e^{-J(\tau)}$, where $Z=\int p(\tau)e^{-J(\tau)}d\tau$ is the normalization constant. The energy vector field $v_\mathcal{E}(\widehat{\tau}^t)$ is defined as:
    \begin{equation}
        v_\mathcal{E}(\widehat{\tau}^t)=v_\mathcal{C}(\widehat{\tau}^t)+\mathcal{G}(\widehat{\tau}^t)
    \end{equation}
    with
    \begin{equation}
        \mathcal{G}(\widehat{\tau}^t)=(\beta\frac{e^{-J(\tau^1)}}{Z(\widehat{\tau}^t)}-1)v(\widehat{\tau}^t),
    \end{equation}
    being value guidance vector field, where $\beta$ is a scaling factor.
\end{definition}
The learning objective of the energy vector field is:
\begin{equation}\label{eq20}
    \begin{aligned}
        \mathcal{L}_{\mathcal{G}}&=\|\mathcal{G}_\psi(\widehat{\tau}^t,t)-(\frac{e^{-J(\tau^1)}}{Z_\phi}-1)v(\widehat{\tau}^t)\|^2_2\\
\mathcal{L}_{Z}&=\|Z_\phi(\widehat{\tau}^t,t)-e^{-J(\tau^1)}\|^2_2,
    \end{aligned}
\end{equation}
where $\mathcal{G}$ and $Z$ are constructed by neural networks with parameter $\psi$ and $\phi$, respectively.

\noindent\textbf{Overall Loss}\ \ \ The total loss of CtrlFlow is as follows:
\begin{equation}
    \mathcal{L}_\text{CtrlFlow}=\mathcal{L}_\text{CFM}(\theta)+\mathcal{L}_\text{Control}(\omega)+\zeta(\mathcal{L}_\mathcal{G}+\mathcal{L}_Z),
\end{equation}
where $\zeta$ is a hyper-parameter to scale the loss.

\section{Experiments}\label{experiments}
In this section, we carefully design the experiment to test and mainly focus on two questions:
\textbf{(1) Can CtrlFlow adapt the dynamic trajectory distributions in online RL?} \textbf{(2) Can CtrlFlow generate high-return trajectories?}

\begin{figure}
    \centering
    \begin{subfigure}[b]{0.23\textwidth}
        \includegraphics[width=\textwidth]{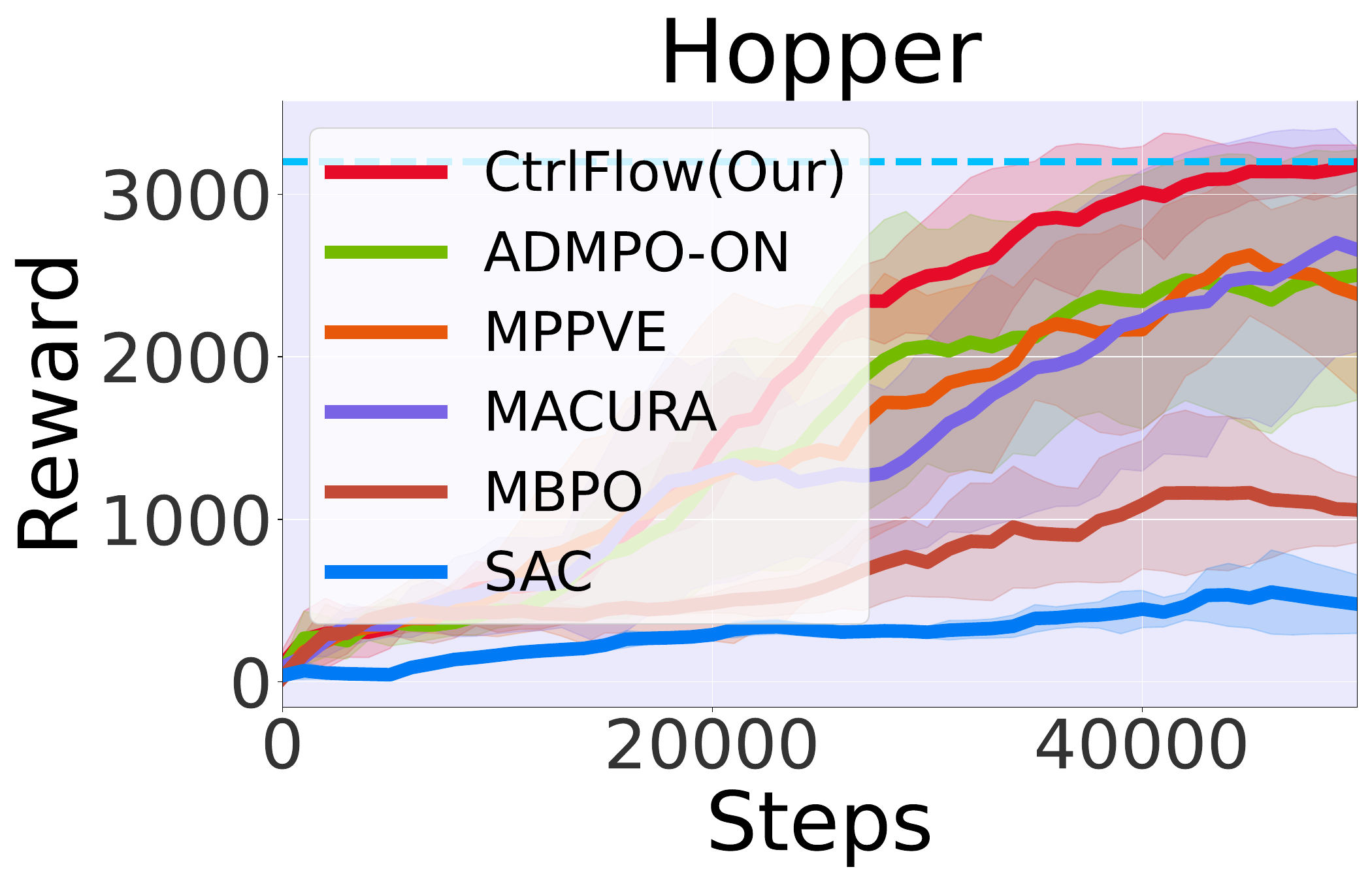}
    \end{subfigure}
    \hfill
    \begin{subfigure}[b]{0.23\textwidth}
        \includegraphics[width=\textwidth]{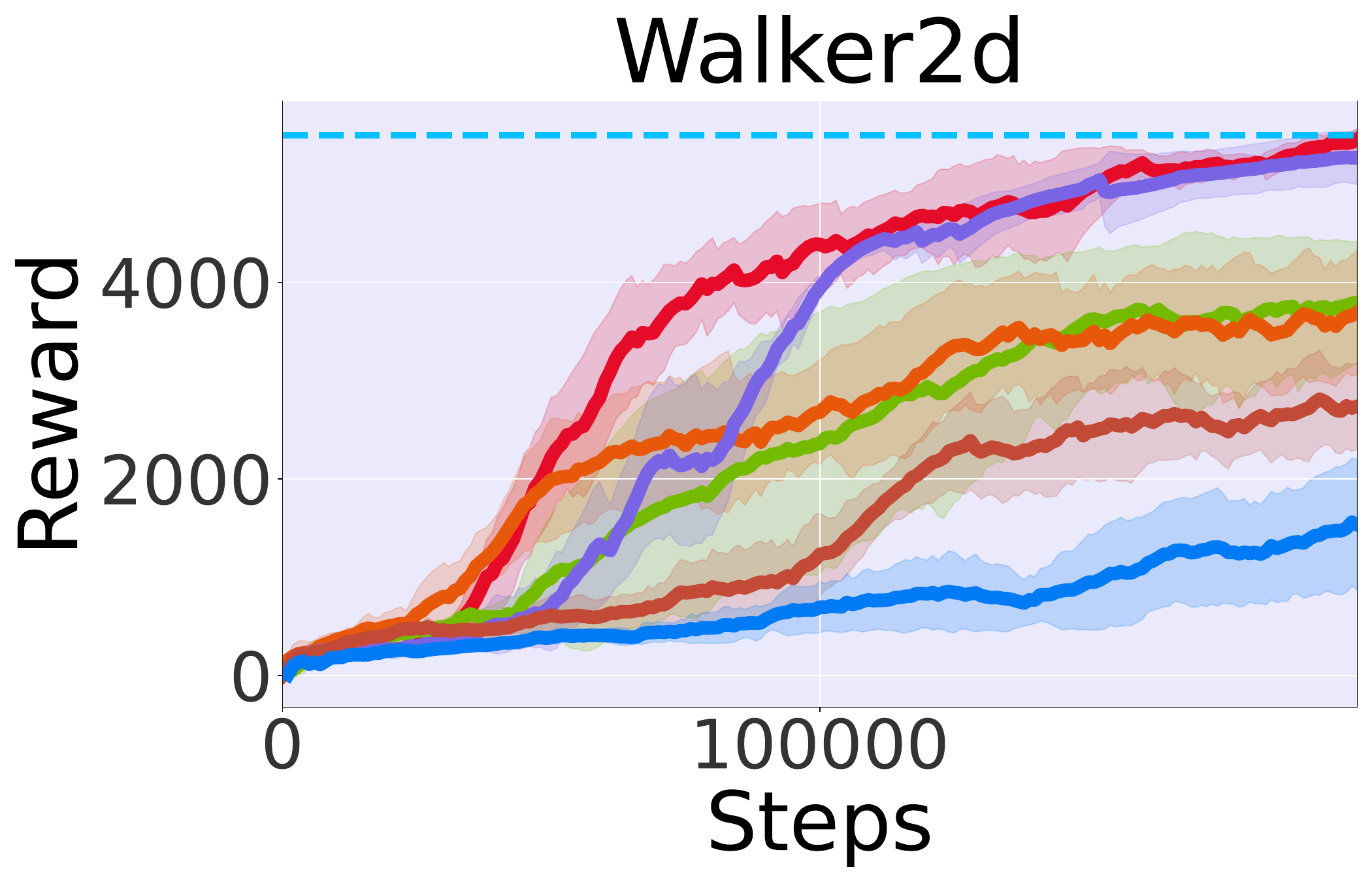}
    \end{subfigure}
    \hfill
    \begin{subfigure}[b]{0.23\textwidth}
        \includegraphics[width=\textwidth]{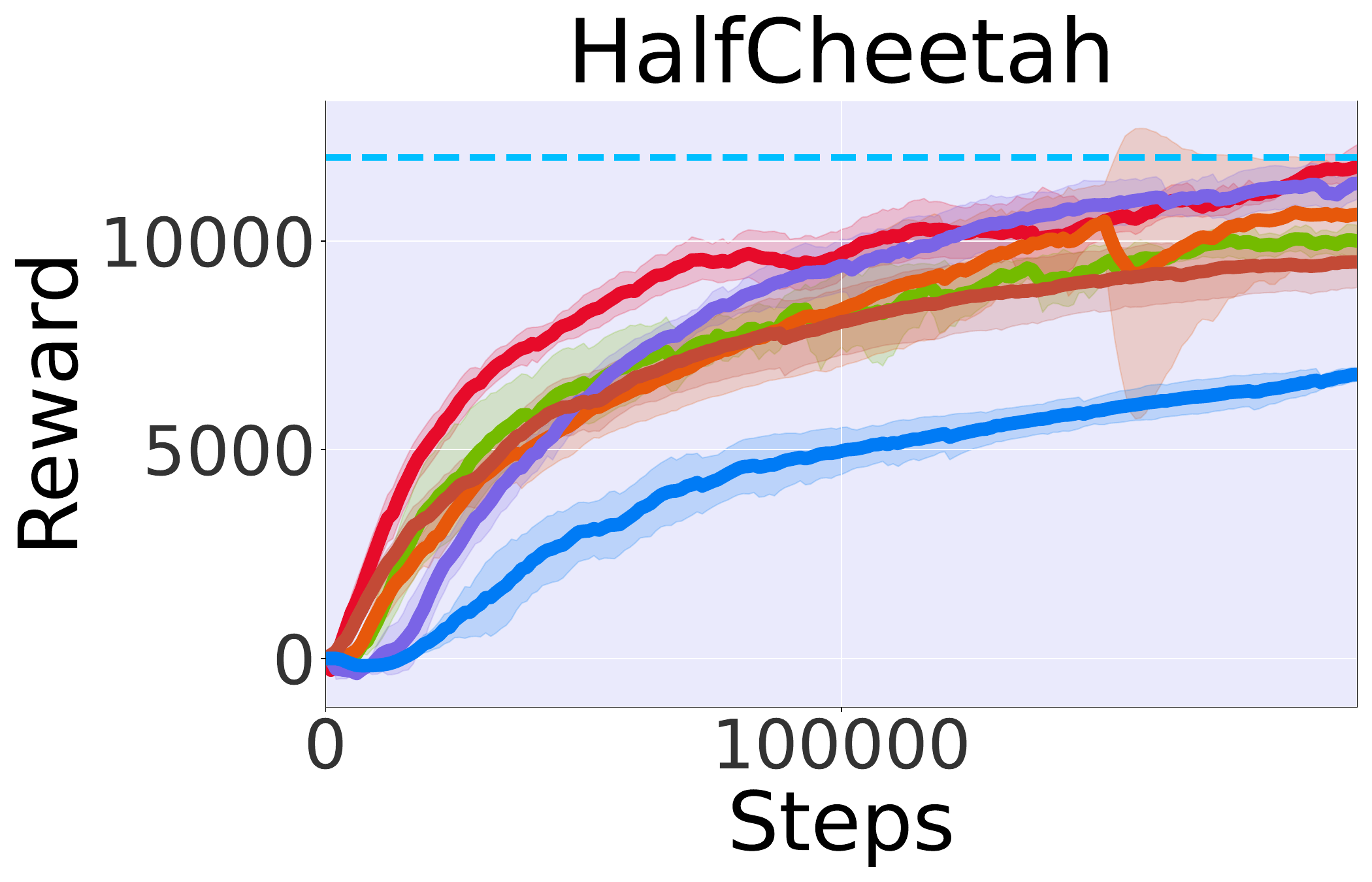}
    \end{subfigure}
    \hfill
    \begin{subfigure}[b]{0.23\textwidth}
        \includegraphics[width=\textwidth]{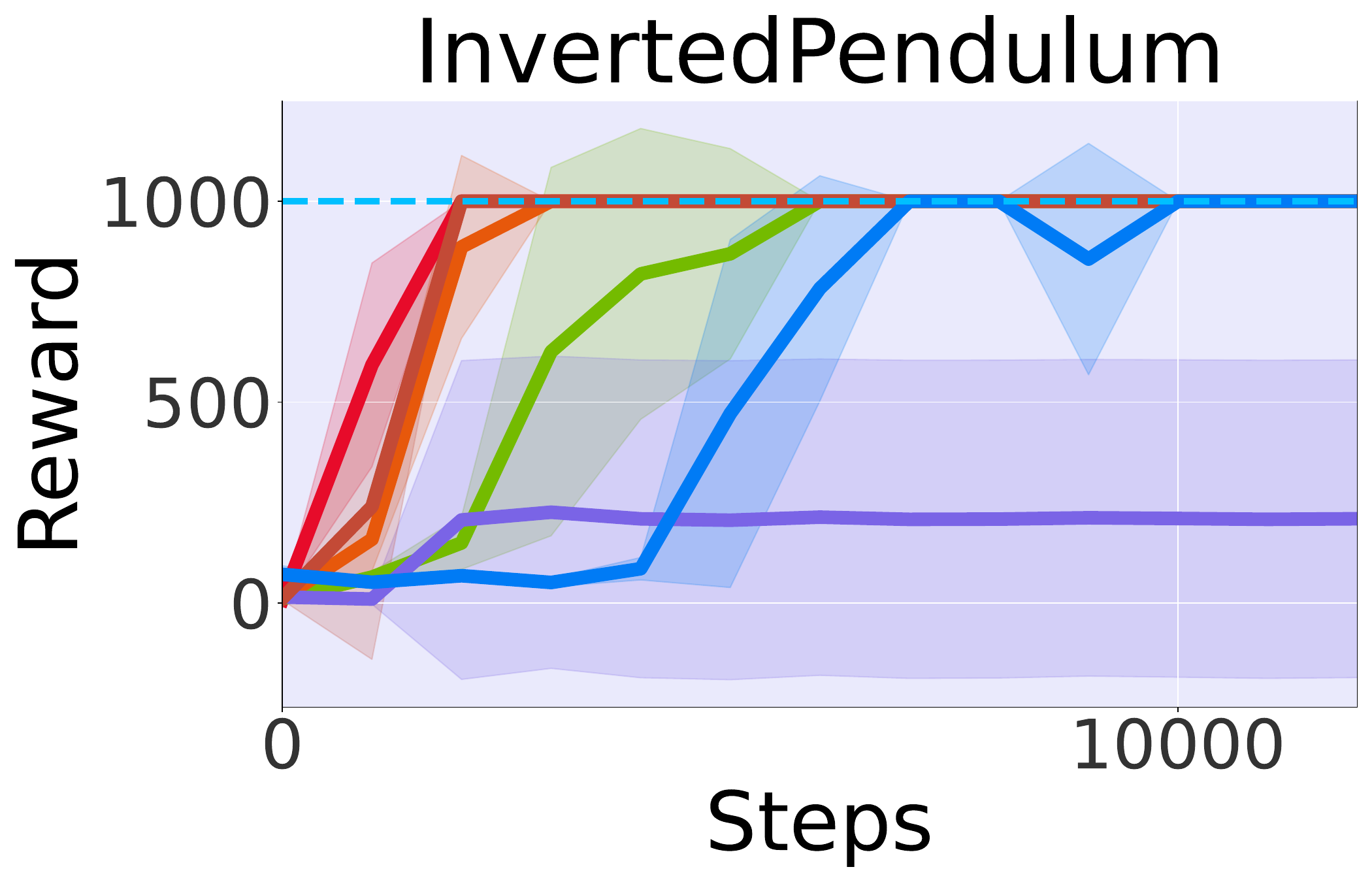}
    \end{subfigure}
    \caption{Performance on the MuJoCo Benchmark. CtrlFlow (red) and other five baselines on control tasks. The blue dashed lines indicate the asymptotic performance of SAC for reference. The solid lines indicate the mean while the shaded areas indicate the standard error over five different seeds.}
    \label{fig:comparison}
\end{figure}

\noindent \textbf{Experimental Setup}\ \ \ We evaluate CtrlFlow on four MoJoCo \cite{MoJoCo} continuous control tasks, including Hopper, Walker2d, HalfCheetah and InvertedPendulum. Hopper, Walker2d, HalfCheetah adopt version v3 and InvertedPendulum adopt version v4, all tasks follow the default settings implemented under the Pytorch framework and run on NVIDIA 3090 GPU. Four model-based methods and one model-free method are selected as our baselines. These include SAC \cite{SAC}, which is the state-of-the-art model-free RL algorithm; ADMPO-ON \cite{ADMPO}, which is the state-of-the-art model-based RL method; MACURA \cite{MACURA}, which uses inherent model uncertainty to consider local accuracy to make rollouts; MPPVE \cite{MPPVE}, which directly computes multi-step policy gradients via plan value estimation when the policy plans start from real states; MBPO \cite{MBPO}, which updates the policy with a mixture of real environmental samples and branched rollouts data. See Appendix C for details about environment.

\noindent \textbf{Main Results}\ \ \ The results on the MuJoCo benchmark are depicted in Figure \ref{fig:comparison}. CtrlFlow achieves great performance after fewer environmental samples than baselines. Take Hopper as an example, CtrlFlow reaches 90\% on its peak performance (around 3000 reward) within only 35k steps, whereas model-based methods, such as MPPVE and ADMPO-ON, require approximately 70k and 60k steps respectively to reach similar reward levels. Model-free method SAC performs significantly worse, achieving only around 1000 after 50k steps. In terms of learning speed, CtrlFlow is 2× faster than ADMPO-ON, 4× faster than MPPVE, and shows an order-of-magnitude improvement over SAC and MACURA. Furthermore, CtrlFlow not only exhibits superior sample efficiency, but also achieves the highest asymptotic performance among all compared methods, stabilizing above 3300 reward.
\begin{figure}[t]
    \centering
    \begin{subfigure}[b]{0.23\textwidth}
        \includegraphics[width=\textwidth]{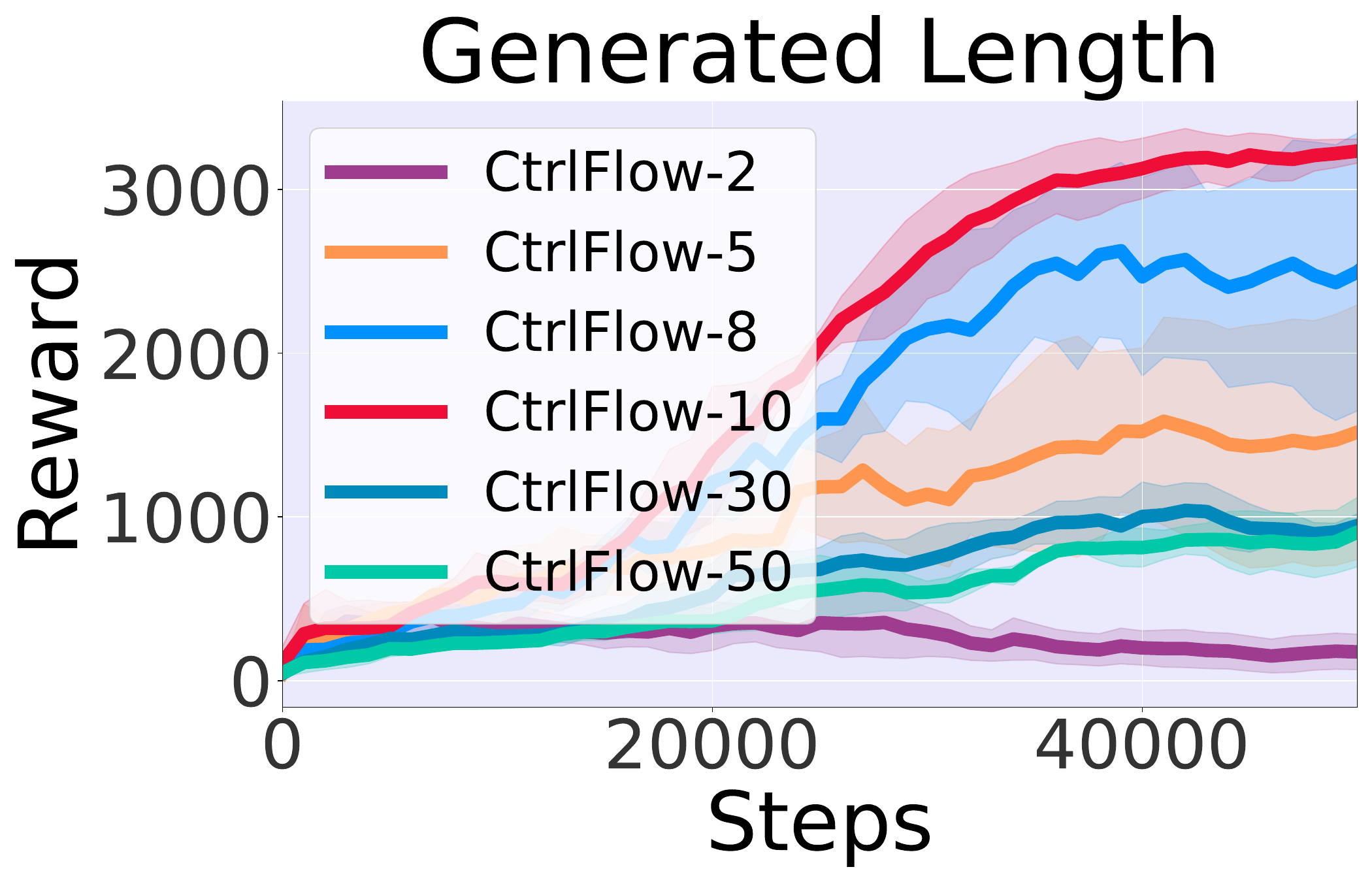}
    \end{subfigure}
    \begin{subfigure}[b]{0.23\textwidth}
        \includegraphics[width=\textwidth]{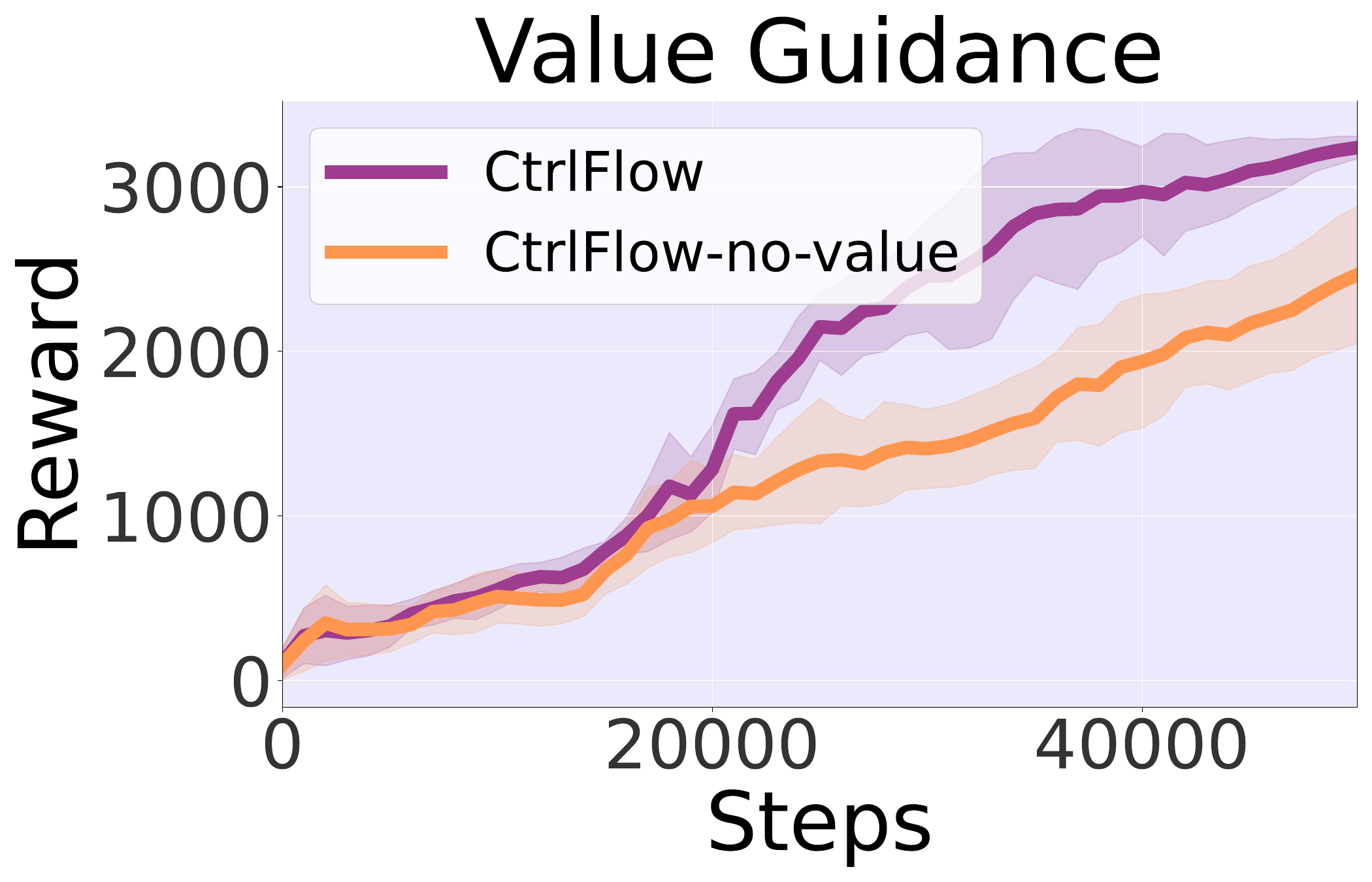}
    \end{subfigure}
    \caption{Study on generated length and value guidance on MuJoCo Hopper-v3 task with five random seeds in 50k steps. (\textbf{Left}) Compare the performance of six different lengths with $h=2,5,8,10,30,50$ by model generation. (\textbf{Right}) The model with and without the value guidance vector field $\mathcal{G}$.}
    \label{fig:study1}
\end{figure}

\noindent \textbf{Ablation on Generated Length}\ \ \ To evaluate algorithm performance across varying trajectory lengths, we compared six generation lengths ($h=2,5,8,10,30,50$) on the MuJoCo Hopper-v3 task (Figure \ref{fig:study1} Left). We design different trajectory lengths to evaluate the model’s performance on short ($h<10$) and long ($h\geq10$) trajectory generation. Results show that CtrlFlow performs better with longer short-horizon trajectories, as they provide richer transition information for policy optimization. However, when the length exceeds 10 (e.g., $h=30$ or $h=50$), performance declines. We attribute this to increased integration complexity and energy demand in long-horizon flow matching, which makes it harder to maintain stable flow constraints and thus degrades policy performance.

\begin{figure}
    \centering
    \begin{subfigure}[b]{0.23\textwidth}
        \includegraphics[width=\textwidth]{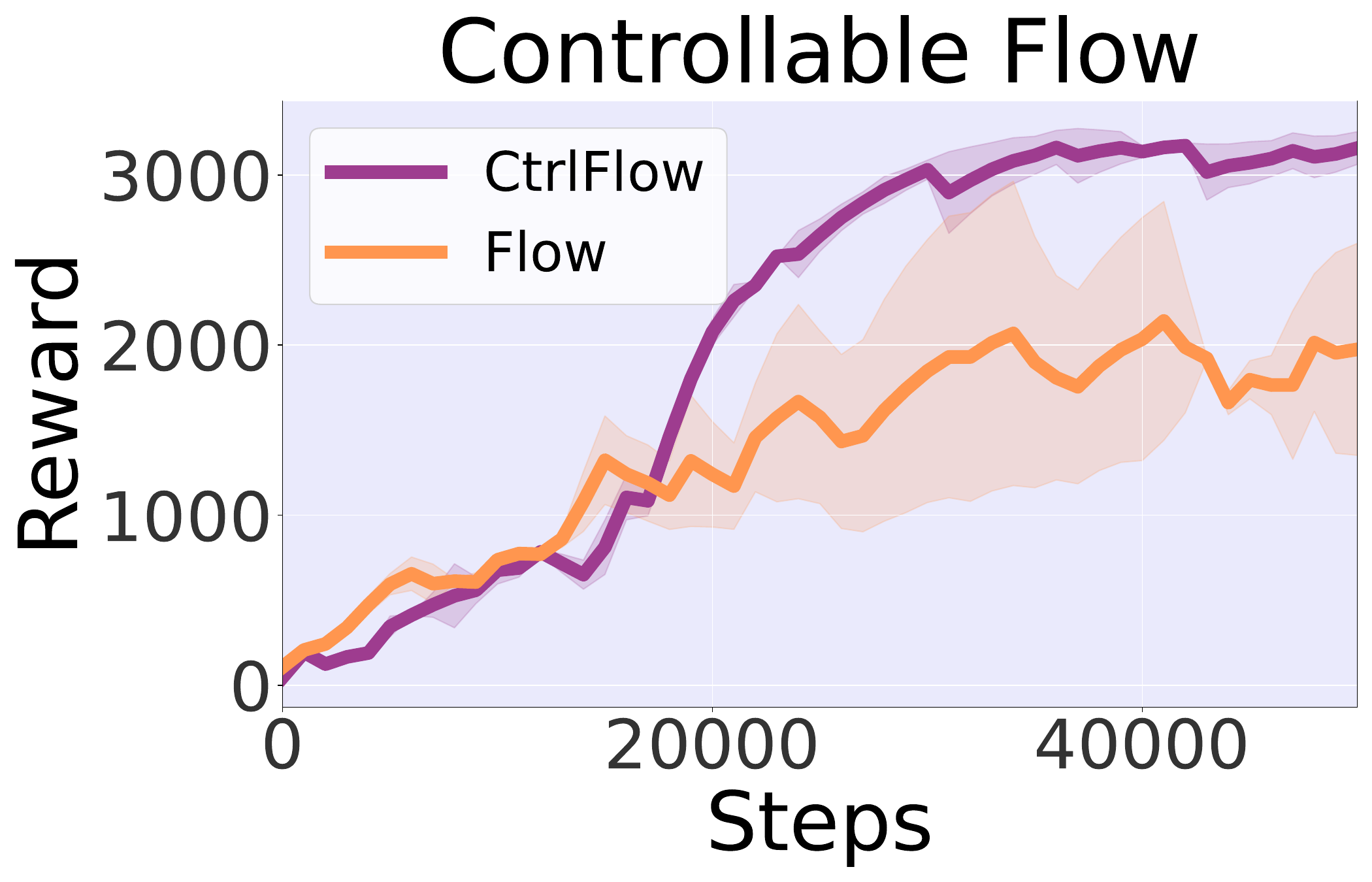}
    \end{subfigure}
    \begin{subfigure}[b]{0.23\textwidth}
        \includegraphics[width=\textwidth]{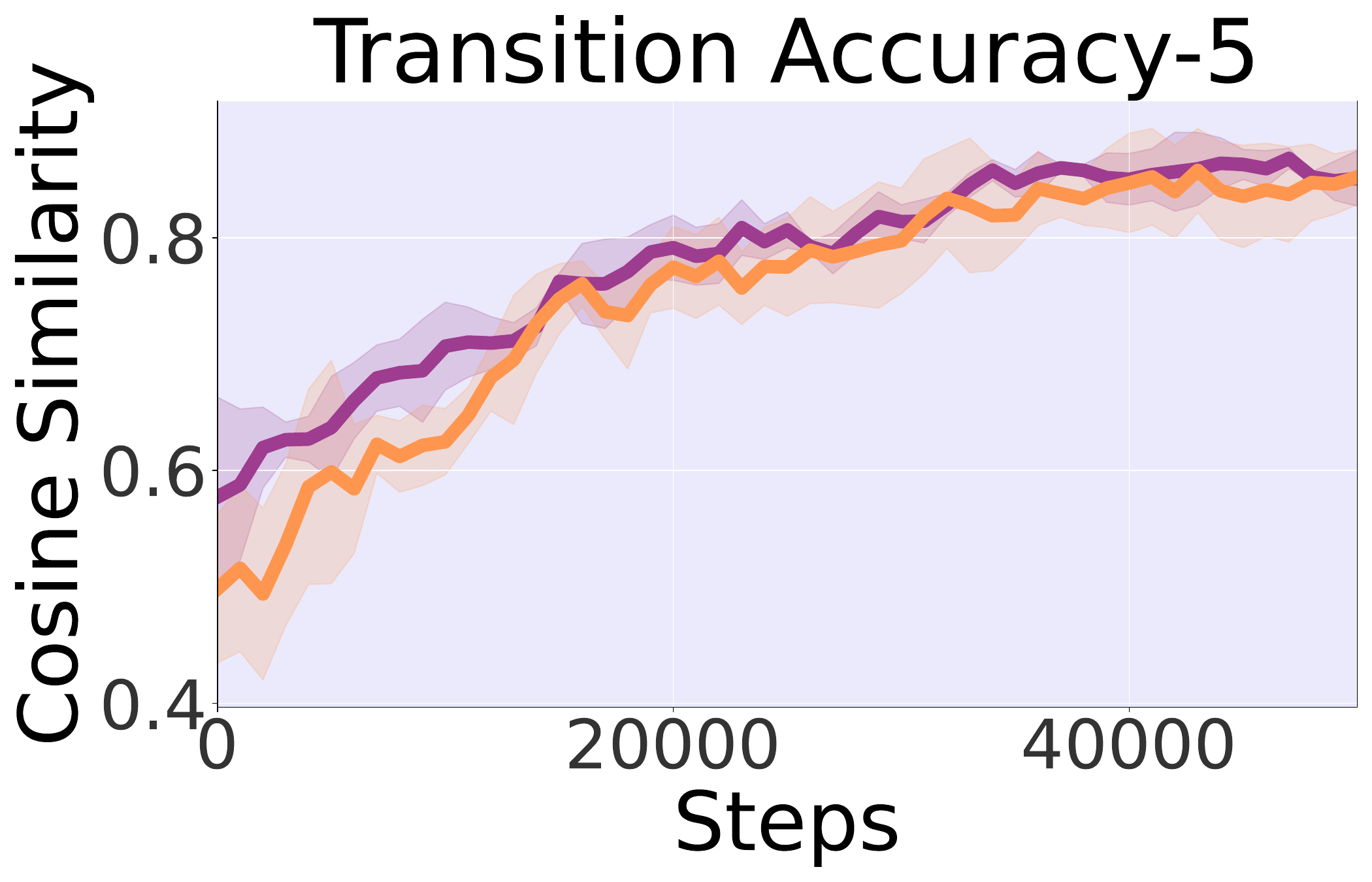}
    \end{subfigure}
    \begin{subfigure}[b]{0.23\textwidth}
        \includegraphics[width=\textwidth]{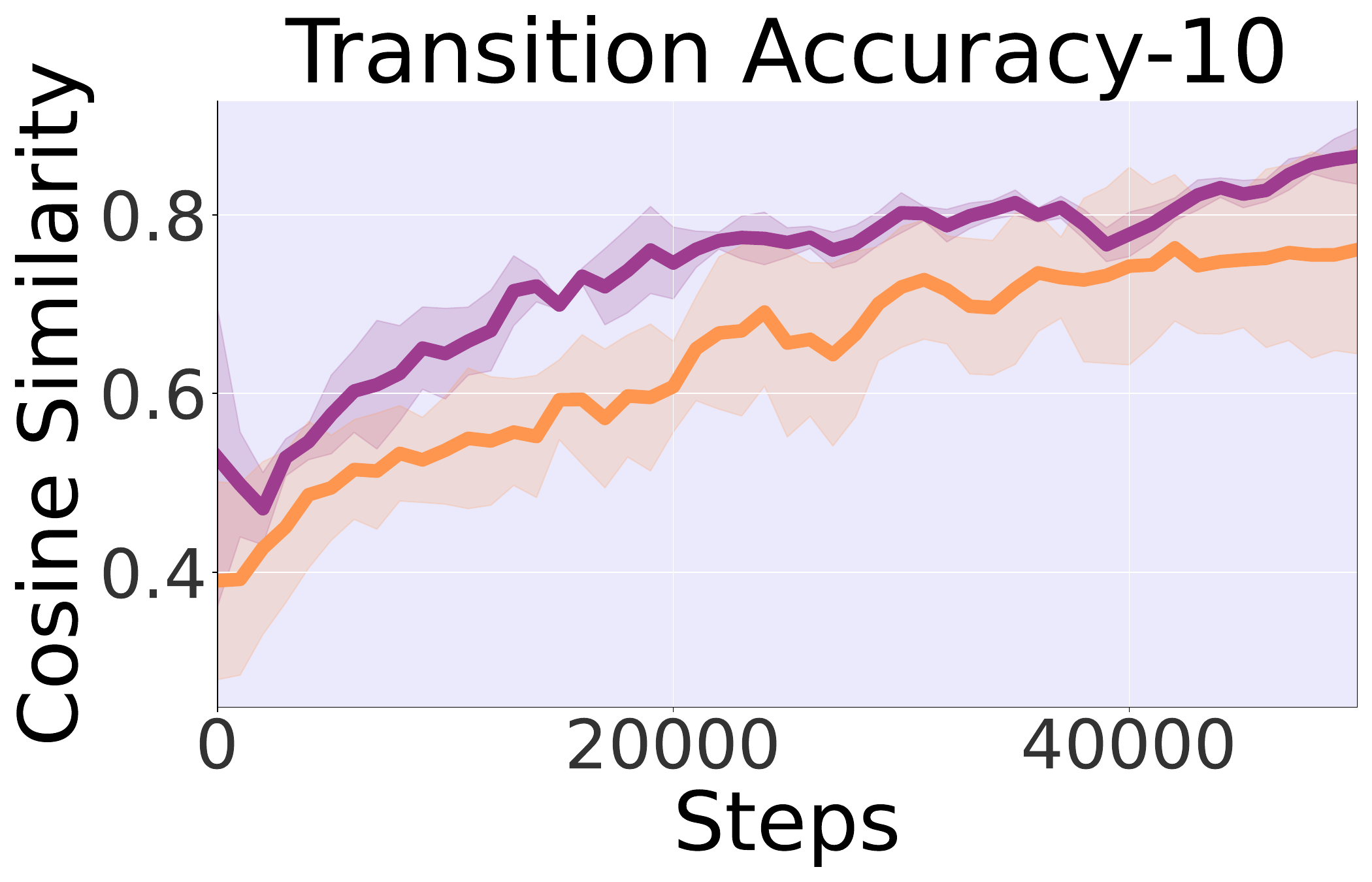}
    \end{subfigure}
    \begin{subfigure}[b]{0.23\textwidth}
        \includegraphics[width=\textwidth]{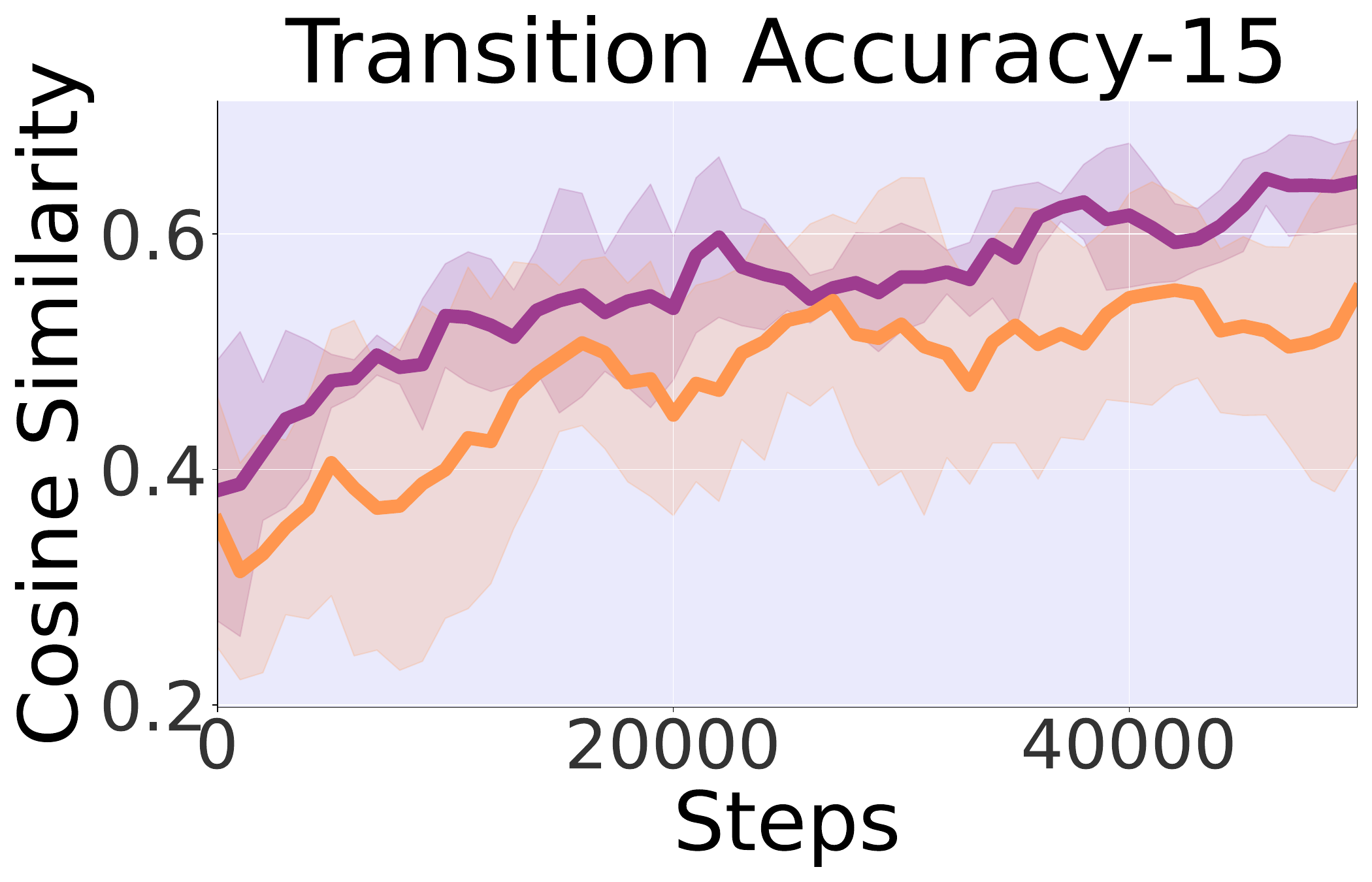}
    \end{subfigure}
    \caption{(\textbf{Top Left}) Study on control theory on MoJoCo Hopper-v3 task in 50k steps. (\textbf{Others}) The comparison of controllable and cosine similarity between the generated final state and the ground truth.}
    \label{fig:study2}
\end{figure}

\noindent \textbf{Ablation on the Importance of Control}\ \ \ An ablation study (Figure \ref{fig:study2} Top left) compares the full CtrlFlow algorithm (purple) with Flow (yellow, control inputs removed) on MuJoCo Hopper-v3 (five random seeds). Results show that CtrlFlow achieves superior convergence and more stable rewards, surpassing Flow after 18k steps due to its controllability-aware sampling via the Controllability Gramian Matrix. In contrast, Flow exhibits unstable rewards between 18k–40k steps, struggling to adapt to shifting data distributions, and fails to converge by 50k steps. To validate state transition dynamics (Figure \ref{fig:study2} Others), we measure cosine similarity between generated and real trajectories from identical initial states. At short horizons (h=5), both methods perform similarly. However, for longer horizons (h=10,15), CtrlFlow maintains significantly higher precision, demonstrating that the Controllability Gramian Matrix effectively minimizes cumulative sampling errors.

\noindent \textbf{Ablation on Value Guidance}\ \ \ In the right of Figure \ref{fig:study1} validate the role of the value-guided energy vector field in the algorithm. CtrlFlow-no-value (without this guidance) serves as the baseline, evaluated on Hopper-v3 with five random seeds. Results indicate that CtrlFlow achieves significantly accelerated convergence, surpassing CtrlFlow-no-value after 20k steps. This reveals its critical role in enabling CtrlFlow has faster convergence relative to MBRL methods. \textbf{Notably, while prior works utilize energy-based methods to accelerate convergence, they exclusively focus on offline RL settings; CtrlFlow represents the first study to validate the efficacy of this approach for data augmentation in online settings.}
\begin{figure}[t]
    \centering
    \begin{subfigure}[b]{0.23\textwidth}
        \includegraphics[width=\textwidth]{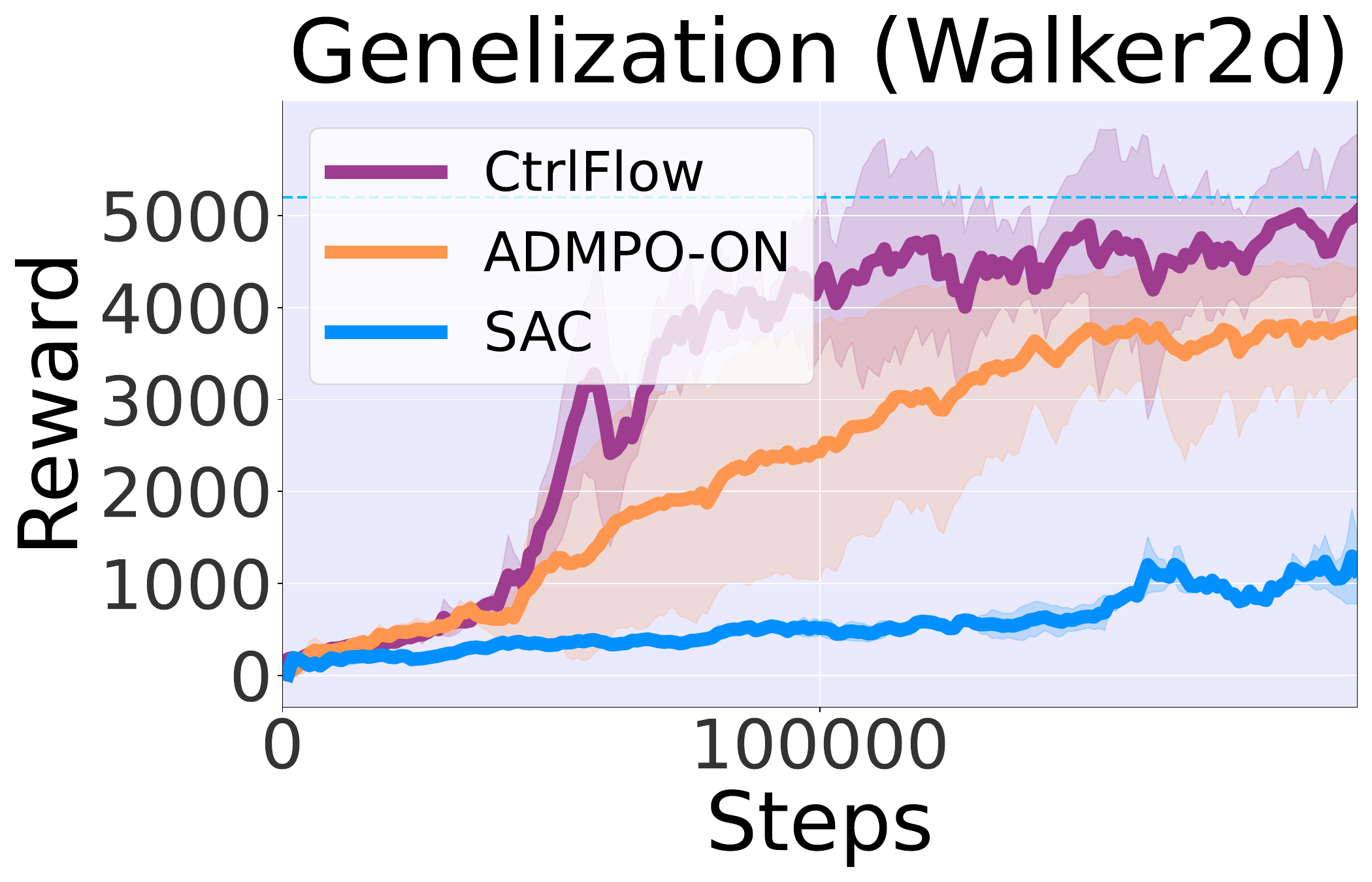}
    \end{subfigure}
    \begin{subfigure}[b]{0.23\textwidth}
        \includegraphics[width=\textwidth]{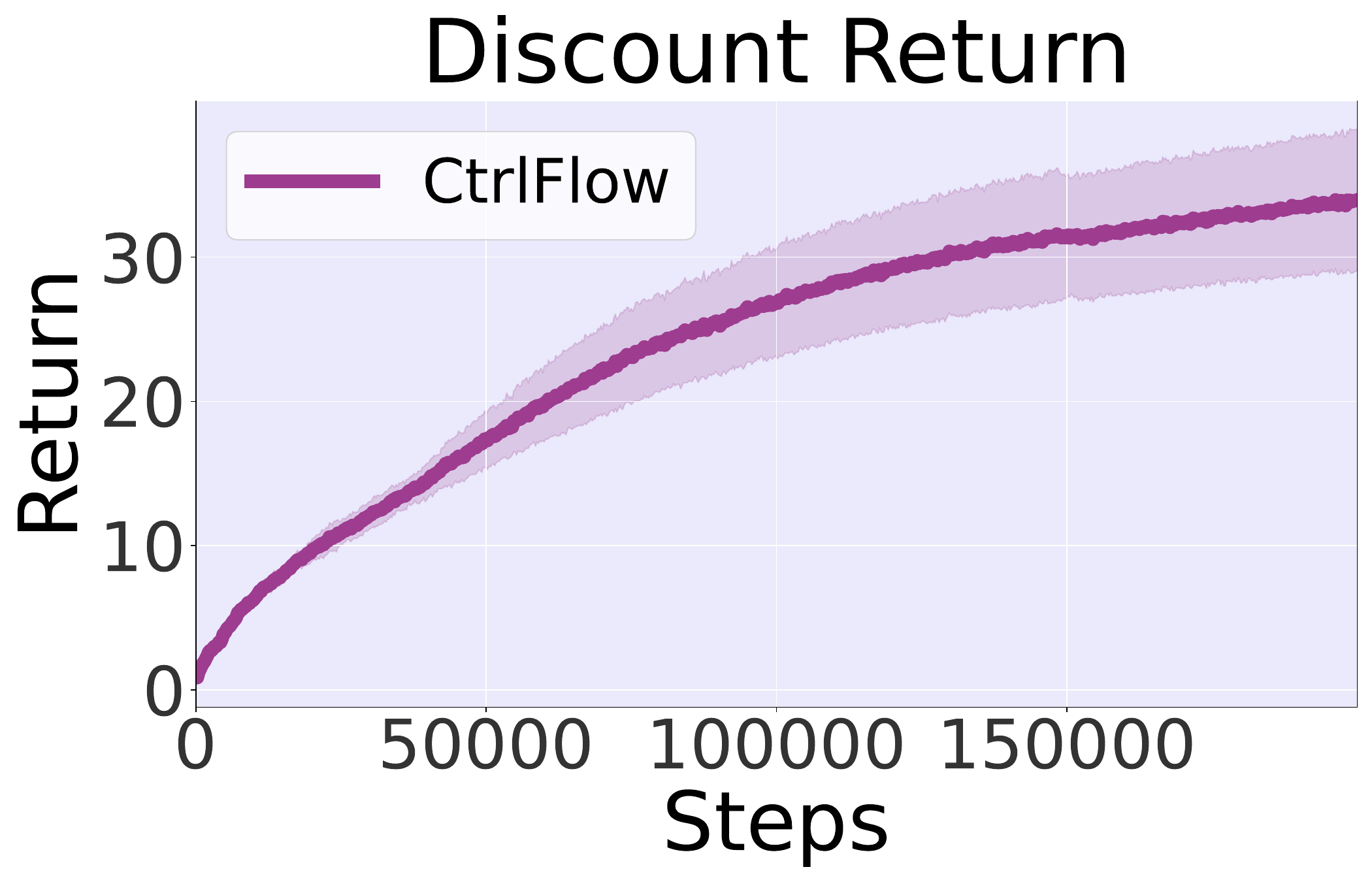}
    \end{subfigure}
    \caption{(\textbf{Left}) Evaluation of generalization ability from HalfCheetah to Walker2d. (\textbf{Right}) Discount Return in Walker2d task generated by model.}
    \label{fig:generation}
\end{figure}

\noindent \textbf{Model Generalization}\ \ \ We demonstrate CtrlFlow's effectiveness as a cross-domain trajectory generator for reinforcement learning. We first train the model in HalfCheetah, then transfer it to Walker2d. Compared to baselines (ADMPO-ON and direct SAC training), CtrlFlow shows superior sample efficiency and faster early convergence (Figure \ref{fig:generation}), indicating successful capture of both task-specific patterns and transferable structural knowledge. The model particularly excels when transferring between tasks with similar state-action representations. This also suggests CtrlFlow learns fundamental movement primitives that generalize across related environments. The strong performance in Walker2d despite being trained on HalfCheetah highlights its robustness to morphological differences.

\section{Conclusion}
In this work, we propose CtrlFlow, the first online RL method generating trajectory-level data via continuous normalizing flows. Unlike model-based approaches, CtrlFlow directly models trajectory distributions, avoiding cumulative errors while enhancing long-term planning and noise robustness. It employs Controllability Gramian Matrix to ensure global controllability during sampling, minimizing control energy for reliable trajectory generation. Value-guided generation via energy vector fields improves distribution alignment, enabling faster convergence than model-based methods. However, CtrlFlow shows limited performance in partially observable environments, where incomplete observations make trajectory modeling difficult. Future work will explore the modeling ability in partially observable settings.

\bibliography{main}

\end{document}